\documentclass{article}

\usepackage{emnlp2016}

\emnlpfinalcopy

\usepackage{subcaption}
\usepackage{amsmath}
\usepackage{multirow}

\usepackage[utf8]{inputenc} 
\usepackage[T1]{fontenc}    
\usepackage{hyperref}       
\usepackage{url}            
\usepackage{booktabs}       
\usepackage{amsfonts}       
\usepackage{nicefrac}       
\usepackage{microtype}      
\usepackage{tikz}
\usetikzlibrary{fit,positioning}
\usepackage{epstopdf}

\title{Sex, drugs, and violence} 

\author{
  Stefania Raimondo \\
  Department of Computer Science\\
  University of Toronto\\
  Toronto, CA\\
  \texttt{sraimond@cs.toronto.edu} 
  \And
  Frank Rudzicz \\
  Toronto Rehabilitation Institute-UHN, and\\
  Department of Computer Science\\
  University of Toronto\\
  Toronto, CA\\
  \texttt{frank@cs.toronto.edu} 
}

\begin{document}

\maketitle


\begin{abstract}
Automatically detecting inappropriate content can be a difficult NLP task, requiring understanding context and innuendo, not just identifying specific keywords. Due to the large quantity of online user-generated content, automatic detection is becoming increasingly necessary. We take a {\em largely} unsupervised approach using a large corpus of narratives from a community-based self-publishing website and a small segment of crowd-sourced annotations. We explore topic modelling using latent Dirichlet allocation (and a variation), and use these to regress appropriateness ratings, effectively automating rating for suitability. The results suggest that certain topics inferred may be useful in detecting {\em latent} inappropriateness -- yielding recall up to 96\% and low regression errors. 
\end{abstract}

\section{Introduction}
The internet has precipitated an explosion of shared user-generated content, with unrestrained access to edit boxes from barely vetted online community members. While this freedom is a boon to many websites, it can lead to abuse if content is inappropriate to the context or viewership. Unfortunately, massive volume makes manual moderation impractical.

If inappropriate content is to be flagged or censored automatically, identifying keywords may be insufficient. Neither innuendo, ever changing slang, nor misspellings can be detected by finite lists of words, which themselves may vary in offensiveness depending on context. A further complication is the paucity of {\em annotated} inappropriate data. In this work, we present a large corpus of narrative content published in the online community Wattpad, augmented by crowd-sourced annotations. While those annotations are {\em relatively} sparse, they provide an opportunity to explore this domain using both unsupervised and partially-supervised methods. 

This paper provides a preliminary exploration of these data using latent Dirichlet allocation (LDA) and an extension, partially-labeled Dirichlet allocation (PLDA), in order to take advantage of the newly acquired labels. Specifically, the texts are classified according to their degree of inappropriateness in three separate categories: sex, violence, and substance abuse. Our goals are to determine a)  whether LDA can be adapted for inappropriateness detection on narrative data, and b) whether PLDA can effectively make use of sparse annotations in this domain.

\section{Background and related work}
Little work has been done on automatically detecting inappropriate or offensive content. 
Thus, major social media websites such as YouTube and Facebook primarily rely on community moderation, in which users ``flag'' inappropriate content which is subsequently dealt with by the organization.
This approach does not guarantee the timely removal of content.

Some existing work aimed to detect short offensive or abusive messages called ``flames''. This included the seminal work by \newcite{spertus_smokey:_1997}, who used a 47-element vector of hand-tuned syntactic and semantic features in a decision tree that achieved $\sim{}$65\% flame-detection and 98\% clean-text detection accuracy. \newcite{razavi_offensive_2010} used an adaptive three-tiered classification scheme using a bag-of-words and a dictionary of abusive/insulting words, phrases and expressions. \newcite{mahmud_detecting_2008} also used keywords and shallow syntactic rules. While flames are a subset of inappropriate content, these methods are unable to detect {\em mature} content, which is more related to subject matter and not restricted to short messages. \newcite{chen_detecting_2012} and \newcite{xu_filtering_2010} also use hand-engineered syntactic features and keywords to detect a limited category of \textit{offensive} passages, e.g., by defining offensive text as containing {\em lexical} pejoratives, profanities, or obscenities. Using message-level lexicosyntactic features extracted from relatively short YouTube comments, \newcite{chen_detecting_2012} obtained at most 98.2\% precision and 94.3\% recall, beating $N$-gram models, although this depended on predetermined measures of word-level `offensiveness' and no latent semantics capable of capturing innuendo.

While research in this area has been minimal, it falls under the larger umbrella of higher-level unsupervised textual analysis of sentiment, opinion, and subjectivity~\cite{liu_sentiment_2010,razavi_offensive_2010}. Unsupervised methods often rely on word contexts and include methods of determining word embeddings and on topic modelling \cite{blei:latent:2003}. 
\newcite{xiang_detecting_2012} used a combination of topic modelling and keyword features as input in a few supervised offensiveness classifiers (e.g., SVN, random forest). 
They `bootstrapped' a set of labelled tweets from known `offensive users' and seed words and found a 5.4\% increase in F1 over the baseline when the number of topics was increased to 50, albeit using only short tweets, and focusing on `bad words' rather than inappropriate {\em content}. 


\section{Data}\label{sec:Data}
Narrative data were shared by Wattpad\footnote{https://www.wattpad.com/}, which is a popular site for social storytelling. These data consist of excerpts from short stories written by members of the community, partitioned into a {\bf Basic} subset from Wattpad's main website ($2,129,156$ excerpts, avg. 931 words/excerpt), and an {\bf AfterDark} subset\footnote{https://www.wattpad.com/afterdark/} intended for ages 17+, and more mature audiences ($47,407$ excerpts, avg. 1597 words/excerpt). For the purposes of this initial analysis, approximately 21,000 excerpts are randomly selected from the former and 10,400 from the latter. These two subsets are then each split equally into training and test sets.




Additionally, over 1000 excerpts were each scored by three annotators for mature content using the CrowdFlower crowd-sourcing service\footnote{https://www.crowdflower.com/}. Annotators were asked to identify specific segments within texts which contained ``inappropriate'' or ``mature'' content, along each of three dimensions (sex, substance abuse, and violence) according to the following 4-point scale:
\begin{enumerate}
\setlength{\itemsep}{-3pt}
   \item  {\bf None}: No inappropriate content. 
   \item {\bf PG}:  Mildly suggestive content or language. Not suitable for children or a professional setting.
   \item {\bf Mature}:	Content appropriate only for a mature audience, as in an R-rated movie. 
    \item {\bf Adult/XXX}: Segment contains explicit, graphic, or disturbing content. 
\end{enumerate}

%

Annotated excerpts were similarly split into training and test sets, with 1/3 of the data comprising the latter. The distribution of ratings in the training and test sets is provided in Table \ref{crowdflower-rating-distribution-table}. Additionally, 289 texts in training and 127 in testing were deemed to be completely appropriate for all audiences. 
Fleiss' inter-annotator agreement statistic gives $\kappa=0.13,p<0.01$ on sexual content, $\kappa=0.01,p=0.33$ on substance abuse, and $\kappa=0.17,p<0.01$ on violence -- i.e. there was only accidental agreement on the rating of drugs. Pearson's correlation, which is more sensitive to ordinal values than Fleiss, generally agrees, with pairwise annotator agreement $\rho >0.20$ for sex and violence, but $\rho < 0.12$ for substance abuse.

\begin{table}[h]
  \caption{Rating frequencies in annotated data}
  \label{crowdflower-rating-distribution-table}
  \centering
  \small
  \begin{tabular}{r|rr|rr|rr}
    \toprule
    & \multicolumn{2}{c}{Sex}&\multicolumn{2}{c}{Drugs}&\multicolumn{2}{c}{Violence}\\
    Rating& Train & Test & Train & Test & Train & Test \\
    \midrule
     2 & 163& 80 & 56 & 43 & 203 & 131\\
    3&   81& 81& 31& 19& 70 & 57\\
   4 & 	12& 9& 3& 8	& 14& 15 \\
     \bottomrule
  \end{tabular}
\end{table}

\section{Methods}
Latent Dirichlet allocation (LDA) is a generative probabilistic model which represents each document as a finite mixture over a set of topics \cite{blei:latent:2003}. 
The only observed variables in our model are the words, $w_{ij}$, which can be inferred using Gibbs sampling and expectation propagation \cite{blei:latent:2003}, associated with inferred weightings on {\em topics}. The number of topics and smoothing hyper-parameters are empirically selected.  


Partially-labeled Dirichlet allocation (PLDA) incorporates per-document labels into the topic model of LDA \cite{ramage_partially_2011}. PLDA assumes that each label is associated with some number of topics and that each topic is associated with a single label. Thus, PLDA is appropriate for applications in which labels are comprehensive and indicate distinct content. 
Like LDA, it associates individual words in a document with topics. While PLDA uses labeled data, topics are still attributed using the same unsupervised approach. 
It also differs from LDA in that each topic is considered to be an amalgamation of sub-topics. 


\section{Experiments}

First, non-English texts are removed and text is tokenized using the Penn Treebank 3 tokenizer\footnote{http://nlp.stanford.edu/software/tokenizer.shtml}. Punctuation is ignored along with 587 English stop words (e.g., pronouns, conjunctions). 
Proper names can appear in many of the learned topics, since they are important ``subject'' words. We also discard documents with fewer than 20 words, and consider only tokens which appear in $D$ documents, where $D$ is selected for each experiment empirically.
LDA and PLDA were performed using Stanford's topic-modelling toolbox\footnote{http://nlp.stanford.edu/software/tmt/tmt-0.4/}. This implementation does not optimize the term and topic smoothing hyper-parameters, which were both set to 0.01 as suggested by the authors of the toolbox.

\subsection{LDA on the unannotated corpus} \label{LDA on Wattpad Corpus}
\begin{figure}[h]
  \centering
  \includegraphics[width=0.5\textwidth]{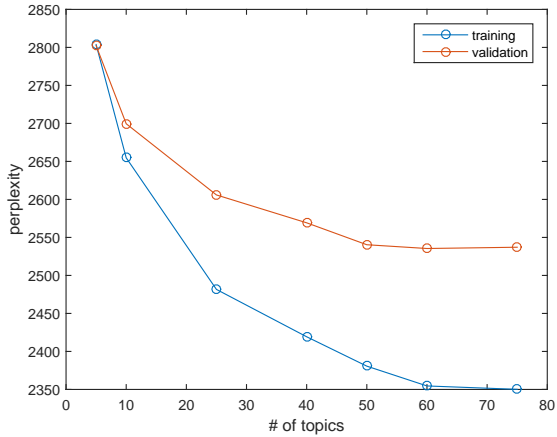}
  \caption{Perplexity of LDA models on Wattpad training and validation data.}
  \label{fig:basic-lda-perplexity}
\end{figure}

In this experiment, LDA is applied to the training sets of {\bf Basic} and {\bf AfterDark} in order to learn a representative  set of combined topics. The intention is to determine whether there is a distinction in topics associated with the two categories, and if this distinction is also present in the ``test'' sets. 
A reasonable way of selecting the number of topics is by using corpus perplexity \cite{blei:latent:2003}. Perplexity is expected to decrease as the number of topics increases. As the model overfits to the training data, the perplexity of the test data is expected to either decrease significantly more slowly or begin to increase; from~\ref{fig:basic-lda-perplexity}, this occurs around $\sim{}$50 topics, so we set $N$ appropriately. Furthermore, the minimum number of documents in which a word may occur is empirically set to 15 in order to reduce proper names from dominating topics in the models with fewer topics. 

Some of the top words associated with each topic are provided in the Supplemental material along with inferred labels. A few observations can be made by `slicing' the output according to the 4-point scale (with the {\bf AfterDark} texts as a fifth point) and considering the per-document topic distribution in each slice. Sixteen topics differ in occurrence by more than 50\% between the topics in the {\bf AfterDark} and {\bf Basic} slices, with six being more prevalent in {\bf AfterDark} including body/sexual (e.g., {\em hips}, {\em hard}), interaction (physical and otherwise, e.g., {\em hands}, {\em feel}), and `crushing' (e.g., {\em moment}, {\em stared}).  



\subsection{PLDA on the annotated corpus}\label{sec:PLDA}

The previous experiment relied on Wattpad's keyword-based rating system, which may not capture latent obscenity. In this section, we provide crowd-sourced annotations 
to PLDA for each text, consisting of the four labels of degree (see section \ref{sec:Data}) provided by the annotators. Following \newcite{miuraTopic}, the `appropriate' label is added to the set for each unlabeled text since annotators did not label them otherwise. This provides texts with explicit `appropriate' labels, similar to using a single latent topic, which can appear in all documents. 

Here, we explore the space for the three hyper-parameters using grid-search, specifically the minimum number of documents per allowable token (min-doc) $\in \{1,5,10,15\}$, the number of background topics (n-bg) $\in \{0,1,5,10\}$, and the number of topics per rating category and level (n-label) $\in \{1,5,10\}$. 

Table \ref{classification-results-table-threshold-05} shows the precision, recall, specificity, and F1 score, averaged over level, for each category, using a threshold of 5\% (there was no significant effect of moving this threshold from 0.1\% to 10\%). Recall is generally in-line with previous work, albeit on a more challenging task of topic modelling, rather than keyword spotting. Across all parameters, precision is <52\%, as labels for inappropriateness tend to be over-applied. This is especially true for the `drugs' category, for which there are the fewest examples. Although not shown, the distribution of error rates over the individual rating levels indicates that  errors arise predominantly from level 4 (i.e., highly inappropriate). Since this analysis treats the labels as nominal, interchangeable classes rather than lying on a continuum, we also perform a regression analysis in the next section.

\begin{table*}[h!]
\centering
\caption{PLDA model classification results for aggregated rating levels, threshold=5\%}
\label{classification-results-table-threshold-05}
\makebox[\linewidth]{
\small
\begin{tabular}{llllllllllllll}
\toprule
         & Bg-topics  			     & 0      &          &        & 1        &         &          & 5      &           &           & 10                                   &        &      \\
         & N-topics                   & 1      & 5      & 10   & 1       & 5       & 10    & 1      & 5        & 10      & 1       & 5      & 10   \\
          \midrule
\multirow{ 4}{*}{sex}     
         & Precision 				& {\textbf{0.36}} & 0.33 & 0.32 & 0.34 & 0.32 & 0.33 & 0.31 & 0.32  & 0.32 & 0.35 & 0.33 & 0.33 \\
         & Recall/Sensitivity  & 0.75 & 0.89 & {\textbf{0.96}} & 0.71 & 0.88 & 0.91 & 0.5 &   0.8    & 0.9 &   0.56 & 0.81 & 0.88 \\
         & Specificity 				 & 0.41 & 0.19 & 0.1   & 0.4    & 0.18 & 0.16 & 0.51 & 0.23 & 0.15 & {\textbf{0.53}} & 0.28 & 0.22 \\
         & F1 				 		      & 0.48 & 0.48 & 0.48 & 0.46 & 0.47 & 0.48 & 0.38 & 0.43 & 0.47 &  0.43 & 0.47 & {\textbf{0.49}} \\
                 \midrule
\multirow{ 4}{*}{violence}   
         & Precision 				& 0.5    & 0.5 & 0.48    & {\textbf{0.52}} & 0.49 & 0.48 & 0.5 &   0.48 &  0.48 & 0.49& 0.48 & 0.49 \\
         & Recall/Sensitivity  & 0.79 & {\textbf{0.94}} & 0.92  & 0.77 & 0.93 & 0.92 & 0.6 &   0.86 &  0.91 & 0.56 & 0.82 & 0.9 \\
         & Specificity 				 & 0.31 & 0.18 & 0.13  & 0.37 & 0.15 & 0.13 & 0.49 & 0.18&   0.16 & {\textbf{0.5}} &   0.25 & 0.2 \\
         & F1 				 		      & 0.61 & {\textbf{0.65}} & 0.63  & 0.62 & 0.64 & 0.63 & 0.55 &  0.61 & 0.63 & 0.52 & 0.61 & 0.63 \\         
                 \midrule
\multirow{ 4}{*}{drugs}      
         & Precision 				& 0.2 & 0.21 & 0.19     & 0.21 & 0.22 &  0.2 &   0.18 &  0.2   &   0.22 & 0.18 & 0.21 & {\textbf{0.23}} \\
         & Recall/Sensitivity  & {\textbf{0.68}} & 0.57 & 0.46  & {\textbf{0.68}} &  0.6 &   0.43 &  0.52 &  0.48 &  0.44 & 0.43 & 0.49 & 0.43 \\
         & Specificity 				 & 0.35 & 0.51 & 0.55  & 0.38 &  0.5 &   0.6  &   0.45 & 0.55 &   0.62 & 0.55 & 0.56 & {\textbf{0.67}} \\
		 & F1 				 		      & 0.31 & 0.31 & 0.27  & {\textbf{0.32}} &  {\textbf{0.32}} & 0.27 &  0.27&  0.28 &  0.29 & 0.25&   0.29& 0.3\\ 
         \bottomrule
\end{tabular}
}
\end{table*}

\subsection{Regression on ratings}

An eventual aim of this work is to automatically provide meaningful ratings for each of sexual, violent, and drug-abuse content, to be used in practice. For this experiment, we trained multilinear regressors for each PLDA model, for each of {\em sex}, {\em drugs}, and {\em violence} separately using support vector machines with stochastic conjugate gradient descent 
Table \ref{tab:regressResults} shows the absolute distance between inferred and average human ratings, for each category. The weighted average multiplies the contribution of each rating error by the inverse of that rating's prevalence in the data; e.g., `appropriate' texts were weighted lower, since there were more of them in the data. The table also shows a `total' row which selects the best PLDA model by averaging the errors for each rating and the average error across rating levels (i.e., each column in the table) for the sex, drugs, and violence regressors trained on that PLDA model. The classification rates are clearly lowest for the extremes of rating levels 1 and 4. Overall, the regression gives absolute errors $<1$ for each of the three categories of content; while promising, future work with more annotated data and non-linear models is recommended.


\begin{table}[h!]
\small
\begin{tabular}{rc | rrrr | r}
\hline
 & $n_{topics}$ & $1$ & $2$& $3$ &$4$&	W.Avg.\\
\hline
sex	 &        	9	&	1.74	& 	 0.69		&0.34		&1.19	&0.77		\\
viol.	&	  20	&	1.55	&	 0.53	 	&0.47		&0.65 &0.73	\\
drugs	&	19	 &     1.42	& 	 0.44   	&0.54		&0.91 &0.83	\\
\hline
Total	&	19 &	1.5 	&0.48	&0.55		&1.34	&0.77	\\
\hline
\hline
\end{tabular}
\caption{Lowest absolute error between inferred and average human ratings, for each category, including associated $n_{topics}$. }
\label{tab:regressResults}
\end{table}

\section{Discussion and Conclusion}
These pilot studies in topic modelling and regression demonstrate promise in the automatic detection of inappropriateness beyond mere keyword spotting. Using unsupervised LDA, as in section \ref{LDA on Wattpad Corpus}, appears to distinguish subject matter that could easily be related to inappropriate content, as indicated by the identification of latent {\em topics}. The {\em degree} to which a topic-based model identifies inappropriate passages missed by simplistic keyword spotting is the subject of future work. As is typical, topics are open to qualitative interpretation, and additional manual validation is required. To a large extent, this work is incomparable to previous work since it is the first in this space to capture latent semantics.

PLDA obviates the need for some manual confirmation, and is largely successful, as it allows for label information to be directly incorporated into the model, which guides topic selection. From the classification results of section \ref{sec:PLDA}, it appears as though models benefit from the ability of PLDA to associate multiple topics with a single label. These models accurately identify inappropriate content in test data for both sex and violence, with some over-estimation. This may partially be due variance in crowd-sourced annotations, as indicated by Fleiss $\kappa$ statistics below 0.2 in all cases (section \ref{sec:Data}).

Further exploration in topic modelling of inappropriate content should aim to reduce false positive rates. The PLDA model may also be incorporated into a larger model that incorporates uncertainty into the provided labels. Expanding the set of annotated data may also yield further improvements, and methods of bootstrapping, potentially starting with the LDA results, may be useful in overcoming the lack of such data. Regardless, the LDA-approach provides a reasonable baseline for further explorations and model development for the automatic detection of inappropriate latent topics in narrative content.

\bibliographystyle{emnlp2016}
\bibliography{project}

\begin{thebibliography}{}

\bibitem[\protect\citename{Blei \bgroup et al.\egroup }2003]{blei:latent:2003}
David~M Blei, Andrew~Y Ng, and Michael~I Jordan.
\newblock 2003.
\newblock Latent dirichlet allocation.
\newblock {\em the Journal of machine Learning research}, 3:993--1022.

\bibitem[\protect\citename{Chen \bgroup et al.\egroup
  }2012]{chen_detecting_2012}
Ying Chen, Yilu Zhou, Sencun Zhu, and Heng Xu.
\newblock 2012.
\newblock Detecting offensive language in social media to protect adolescent
  online safety.
\newblock In {\em Privacy, {Security}, {Risk} and {Trust} ({PASSAT}), 2012
  {International} {Conference} on and 2012 {International} {Confernece} on
  {Social} {Computing} ({SocialCom})}, pages 71--80. IEEE.

\bibitem[\protect\citename{Liu}2010]{liu_sentiment_2010}
Bing Liu.
\newblock 2010.
\newblock Sentiment {Analysis} and {Subjectivity}.
\newblock {\em Handbook of natural language processing}, 2:627--666.

\bibitem[\protect\citename{Mahmud \bgroup et al.\egroup
  }2008]{mahmud_detecting_2008}
Altaf Mahmud, Kazi~Zubair Ahmed, and Mumit Khan.
\newblock 2008.
\newblock Detecting flames and insults in text.
\newblock In {\em Proceedings of the Sixth International Conference on Natural
  Language Processing}.

\bibitem[\protect\citename{Miura \bgroup et al.\egroup }2013]{miuraTopic}
Yasuhide Miura, Keigo Hattori, and Tomoko Ohkuma.
\newblock 2013.
\newblock Topic {Modeling} with {Sentiment} {Clues} and {Relaxed} {Labeling}
  {Schema}.
\newblock In {\em Proceedings of the 3rd Workshop on Sentiment Analysis where
  AI meets Psychology (SAAIP 2013)}.

\bibitem[\protect\citename{Ramage \bgroup et al.\egroup
  }2011]{ramage_partially_2011}
Daniel Ramage, Christopher~D. Manning, and Susan Dumais.
\newblock 2011.
\newblock Partially {Labeled} {Topic} {Models} for {Interpretable} {Text}
  {Mining}.
\newblock In {\em Proceedings of the 17th {ACM} {SIGKDD} {International}
  {Conference} on {Knowledge} {Discovery} and {Data} {Mining}}, {KDD} '11,
  pages 457--465, New York, NY, USA. ACM.

\bibitem[\protect\citename{Razavi \bgroup et al.\egroup
  }2010]{razavi_offensive_2010}
Amir~H Razavi, Diana Inkpen, Sasha Uritsky, and Stan Matwin.
\newblock 2010.
\newblock Offensive language detection using multi-level classification.
\newblock In {\em Advances in {Artificial} {Intelligence}}, pages 16--27.
  Springer.

\bibitem[\protect\citename{Spertus}1997]{spertus_smokey:_1997}
Ellen Spertus.
\newblock 1997.
\newblock Smokey: {Automatic} {Recognition} of {Hostile} {Messages}.
\newblock In {\em Proceedings of the {Fourteenth} {National} {Conference} on
  {Artificial} {Intelligence} and {Ninth} {Conference} on {Innovative}
  {Applications} of {Artificial} {Intelligence}}, {AAAI}'97/{IAAI}'97, pages
  1058--1065, Providence, Rhode Island. AAAI Press.

\bibitem[\protect\citename{Xiang \bgroup et al.\egroup
  }2012]{xiang_detecting_2012}
Guang Xiang, Bin Fan, Ling Wang, Jason Hong, and Carolyn Rose.
\newblock 2012.
\newblock Detecting offensive tweets via topical feature discovery over a large
  scale twitter corpus.
\newblock In {\em Proceedings of the 21st {ACM} international conference on
  {Information} and knowledge management}, pages 1980--1984. ACM.

\bibitem[\protect\citename{Xu and Zhu}2010]{xu_filtering_2010}
Zhi Xu and Sencun Zhu.
\newblock 2010.
\newblock Filtering {Offensive} {Language} in {Online} {Communities} using
  {Grammatical} {Relations}.
\newblock {\em Proceedings of Collaboration, Electronic messaging, Anti-Abuse
  and Spam Conference 2010}.

\end{thebibliography}

\end{document}